# Distributed Deep Q-Learning

Hao Yi Ong[1], Kevin Chavez[1], and Augustus Hong[1]

*Abstract*— We propose a distributed deep learning model to learn control policies directly from high-dimensional sensory input via reinforcement learning. The model is based on the deep Q-network, a convolutional neural network trained with Q-learning. Its input is raw pixels and its output is a value function estimating future rewards from taking an action at a given state. To distribute the deep Q-network training, we adapt the *DistBelief* software framework to train the reinforcement learning agents. As a result, the method is completely asynchronous and scales well with the number of machines. We demonstrate that the deep Q-network agent, receiving only the pixels and the game score as inputs, was able to achieve reasonable success on a simple game with minimal parameter tuning.

## I. INTRODUCTION

Reinforcement learning (RL) agents face a tremendous challenge in optimizing their control of a system approaching real-world complexity: they must derive efficient representations of the environment from high-dimensional sensory inputs and use these to generalize past experience to new situations. While past work in RL has shown that with good hand-crafted features agents are able to learn good control policies, their applicability has been limited to domains where such features have been discovered, or to domains with fully observed, low-dimensional state spaces [1]–[3].

We consider the problem of efficiently scaling a deep learning algorithm to control a complicated system with high-dimensional sensory inputs. The basis of our algorithm is a RL agent called a deep Q-network (DQN) [4], [5] that combines RL with a class of artificial neural networks known as deep neural networks [6]. DQN uses an architecture called the deep convolutional network, which utilizes hierarchical layers of tiled convolutional filters to exploit the local spatial correlations present in images. As a result, this architecture is robust to natural transformations such as changes of viewpoint and scale [7].

In practice, increasing the scale of deep learning with respect to the number of training examples or the number of model parameters can drastically improve the performance of deep neural networks [8], [9]. To train a deep network with many parameters on multiple machines efficiently, we adapt a software framework called *DistBelief* to the context of the training of RL agents [10]. Our new framework supports data parallelism, thereby allowing us to potentially utilize computing clusters with thousands of machines for large-scale distributed training, as shown in [10] in the context of unsupervised image classification. To achieve model parallelism, we use Caffe, a deep learning framework developed for image recognition that distributes training across multiple processor cores [11].

The contributions of this paper are twofold. First, we develop and implement a software framework adapted that supports model and data parallelism for DQN. Second, we demonstrate and analyze the performance of our distributed RL agent. The rest of this paper is organized as follows. Section II introduces the background on the class of machine learning problem our algorithm solves. This is followed by Section III and Section IV, which detail the serial DQN and our approach to distributing the training. Section V discusses our experiments on a classic video game, and some concluding remarks are drawn and future works mentioned in Section VI.

## II. BACKGROUND

We begin with a brief review on MDPs and reinforcement learning (RL).

### A. Markov decision process

In an MDP, an agent chooses action $a_t$ at time $t$ after observing state $s_t$. The agent then receives reward $r_t$, and the state evolves probabilistically based on the current state-action pair. The explicit assumption that the next state only depends on the current state-action pair is referred to as the Markov assumption. An MDP can be defined by the tuple $(\mathcal{S}, \mathcal{A}, T, R)$, where $\mathcal{S}$ and $\mathcal{A}$ are the sets of all possible states and actions, respectively, $T$ is a probabilistic transition function, and $R$ is a reward function. $T$ gives the probability of transitioning into state $s'$ from taking action $a$ at the current state $s$, and is often denoted $T(s, a, s')$. $R$ gives a scalar value indicating the immediate reward received for taking action $a$ at the current state $s$ and is denoted $R(s, a)$.

To solve an MDP, we compute a policy $\pi^\star$ that, if followed, maximizes the expected sum of immediate rewards from any given state. The optimal policy is related to the optimal state-action value function $Q^\star(s, a)$, which is the expected value when starting in state $s$, taking action $a$, and then following actions dictated by $\pi^\star$. Mathematically, it obeys the Bellman recursion

$$Q^\star(s,a) = R(s,a) + \sum_{s' \in \mathcal{S}} T(s,a,s') \max_{a' \in \mathcal{A}} Q^\star(s',a').$$

The state-action value function can be computed using a dynamic programming algorithm called value iteration. To

---

[1]H. Y. Ong, K. Chavez, and A. Hong are with the Departments of Mechanical Engineering, Electrical Engineering, and Computer Science, respectively, at Stanford University, Stanford, CA 94305, USA {haoyi, kjchavez, auhong}@stanford.edu

obtain the optimal policy for state *s*, we compute

$$\pi^\star(s) = \underset{a \in \mathcal{A}}{\operatorname{argmax}} Q^\star(s,a).$$

*B. Reinforcement learning*

The problem reinforcement learning seeks to solve differs from the standard MDP in that the state space and transition and reward functions are unknown to the agent. The goal of the agent is thus to both build an internal representation of the world and select actions that maximizes cumulative future reward. To do this, the agent interacts with an environment through a sequence of observations, actions, and rewards and learns from past experience.

In our algorithm, the deep Q-network builds its internal representation of its environment by explicitly approximating the state-action value function $Q^\star$ via a deep neural network. Here, the basic idea is to estimate

$$Q^\star(s,a) = \max_\pi E[R_t \mid s_t = s, a_t = a, \pi],$$

where $\pi$ maps states to actions (or distributions over actions), with the additional knowledge that the optimal value function obeys Bellman equation

$$Q^\star(s,a) = \underset{s' \sim \mathcal{E}}{E}\left[r + \gamma \max_{a'} Q^\star(s',a') \mid s,a\right],$$

where $\mathcal{E}$ is the MDP environment.

III. APPROACH

This section presents the general approach adapted from the serial deep Q-learning in [4], [5] to our purpose. In particular, we discuss the neural network architecture, the iterative training algorithm, and a mechanism that improves training convergence stability.

*A. Preprocessing and network architecture*

Working directly with raw video game frames can be computationally demanding. Our algorithm applies a basic preprocessing step aimed at reducing the input dimensionality. Here, the raw frames are gray-scaled from their RGB representation and down-sampled to a fixed size for input to the neural network. For this paper, the function applies this preprocessing to the last four frames of a sequence and stacks them to produce the input to the state-action value function $Q$.

We use an architecture in which there is a separate output unit for each possible action, and only the state representation is an input to the neural network; i.e., the preprocessed four frames sequence. The outputs correspond to the predicted Q-values of the individual action for the input size. The main advantage of this type of architecture is the ability to compute Q-values for all possible actions in a given state with only a single forward pass through the network. The exact architecture is presented in Appendix A, but a brief outline is as follows. The neural network takes as input a sequence of four frames preprocessed as described above. The first few layers are convolutional layers that applies a rectifier nonlinearity

$$f(x) = \max(0,x),$$

which was empirically observed to model real/integer valued inputs well [12], [13], as is required in our case. The remaining layers are fully-connected linear layers with a single output for each valid action. The number of valid actions varies with the game application. The neural network is implemented on Caffe [11], which is a versatile deep learning framework that allows us to define the network architecture and training parameters freely. And because Caffe is designed to take advantage of all available computing resources on a machine, we can easily achieve model parallelism using the software.

*B. Q-learning*

We parameterize the approximate value function $Q(s,a \mid \theta)$ using the deep convolutional network as described above, in which $\theta$ are the parameters of the Q-network. These parameters are iteratively updated by the minimizers of the loss function

$$L_i(\theta_i) = \underset{s,a \sim \rho(\cdot)}{E}\left[(y_i - Q(s,a;\theta_i))^2\right], \quad (1)$$

with iteration number $i$, target $y_i = E_{s' \sim \mathcal{E}}[r + \gamma \max_{a'} Q(s',a';\theta_{i-1}) \mid s,a]$, and "behavior distribution" (exploration policy $\rho(s,a)$. The optimizers of the Q-network loss function can be computed by gradient descent

$$Q(s,a;\theta) := Q(s,a;\theta) + \alpha \nabla_\theta Q(s,a;\theta),$$

with learning rate $\alpha$.

For computational expedience, the parameters are updated after every time step; i.e., with every new experience. Our algorithm also avoids computing full expectations, and we train on single samples from $\rho$ and $\mathcal{E}$. This results in the Q-learning update

$$Q(s,a) := Q(s,a) + \alpha \left(r + \gamma \max_{a'} Q(s',a') - Q(s,a)\right).$$

The procedure is an *off-policy* training method [14] that learns the policy $a = \operatorname{argmax}_a Q(s,a;\theta)$ while using an exploration policy or behavior distribution selected by an $\varepsilon$-greedy strategy.

The target network parameters used to compute $y$ in Eq. (1) are only updated with the Q-network parameters every $C$ steps and are held fixed between individual updates. These staggered updates stabilizes the learning process compared to the standard Q-learning process, where an update that increases $Q(s_t, a_t)$ often also increases $Q(s_{t+1}, a)$ for all $a$ and hence also increases the target $y$. These immediate updates could potentially lead to oscillations or even divergence of the policy. Deliberately introducing a delay between the time an update to $Q$ is made and the time the update affects the targets makes divergence or oscillations more unlikely [4], [5].

## C. Experience replay

Reinforcement learning can be unstable or even diverge when a nonlinear function approximator such as a neural network is used to represent the value function [15]. This instability has several causes. A source of instability is the correlations present in the sequence of observations. Additionally, the fact that small updates to $Q$ may significantly change the policy and therefore change the data distribution. Finally, the correlations between $Q$ and its target values can cause the learning to diverge. [4], [5] address these instabilities uses a mechanism called experience replay that randomizes over the data, thereby removing correlations in the observation sequence and smoothing over changes in the data distribution.

In experience replay, the agent's experiences at each time step is stored as a tuple $e_t = (s_t, a_t, r_t, s_{t+1})$ in a dataset $\mathcal{D}_t = (e_1, \ldots, e_t)$ pooled over many game instances (defined by the start and termination of a game) into a replay memory. During the inner loop of the algorithm, we apply Q-learning updates, or minibatch updates, to samples of experience drawn at random from the replay dataset.

This approach demonstrates several improvements over standard Q-learning. First, each step of experience is potentially used in many weight updates, thus allowing for greater data efficiency. Second, learning directly from consecutive samples is inefficient due to the strong correlations between the samples. Randomizing the samples breaks these correlations and reduces the update variance. Last, when learning on-policy the current parameters determine the next data sample that the parameters are trained on. For instance, if the maximizing action is to move left then the training samples will be dominated by samples from the left-hand side; if the maximizing action then changes to the right then the training distribution will also change. Unwanted feedback loops may therefore arise and the method could get stuck in a poor local minimum or even diverge.

With experience replay, the behavior distribution is averaged over many of its states, smoothing out learning and avoiding oscillations or divergence in the parameters. Note that when learning by experience replay, it is necessary to learn off-policy because our current parameters are different to those used to generate the sample, which motivates the choice of Q-learning.

In practice, our algorithm only stores the last $N$ experience tuples in the replay memory. It then samples uniformly at random from $D$ when performing updates. This approach is limited because the memory buffer does not differentiate important transitions and always overwrites with recent transitions owing to the finite memory size $N$. Similarly, the uniform sampling gives equal importance to all transitions in the replay memory. A more sophisticated sampling strategy might emphasize transitions from which we can learn the most, similar to prioritized sweeping [16].

## IV. DISTRIBUTED DEEP Q-LEARNING

Algorithm 1 and Algorithm 2 define the distributed deep Q-learning algorithm. In this section we discuss some important points about parallelism and performance.

**Algorithm 1** Worker $k$: ComputeGradient
**state:** Replay dataset $\mathcal{D}_k$, game state $s_t$, target model $\hat{\theta}$, target model generation $\ell$

---

**Fetch** model $\theta$ and iteration number $n$ from server.
**if** $n \geq \ell C$ **then**
    $\hat{\theta} \leftarrow \theta$
    $\ell \leftarrow \lfloor n/C \rfloor + 1$
$q \leftarrow \max\{(N_{\max} - n)/N_{\max}, 0\}$
$\varepsilon \leftarrow (1-q)\varepsilon_{min} + q\varepsilon_{max}$
**Select** action $a_t = \begin{cases} \max_a Q(\phi(s_t), a; \theta) & \text{w.p. } 1-\varepsilon \\ \text{random action} & \text{otherwise} \end{cases}$
**Execute** action $a_t$ and observe reward $r_t$ and frame $x_{t+1}$
**Append** $s_{t+1} = (s_t, a_t, x_{t+1})$ and preprocess $\phi_{t+1} = \phi(s_{t+1})$
**Store** experience $(\phi_t, a_t, r_t, \phi_{t+1})$ in $\mathcal{D}_k$
**Uniformly sample** minibatch of experiences $X$ from $\mathcal{D}_k$, where $X_j = (\phi_j, a_j, r_j, \phi_{j+1})$
**Set** $y_j = \begin{cases} r_j & \text{if } \phi_{j+1} \text{ terminal} \\ r_j + \gamma \max_{a'} \hat{Q}(\phi_{j+1}, a'; \hat{\theta}) & \text{otherwise} \end{cases}$

$$\Delta\theta \leftarrow \nabla_\theta f(\theta; X) = \frac{1}{b}\sum_{i=1}^{b} \nabla_\theta \left(\frac{1}{2}(Q(\phi_i, a_i; \theta) - y_i)^2\right)$$

**Send** $\Delta\theta$ to parameter server.

---

## A. Data parallelism

The serial Deep Q-learning algorithm uses stochastic gradient descent to train the Q network. SGD is an inherently sequential algorithm, but various attempts have been made to parallelize it effectively. We implement a variant of Downpour SGD that takes advantage of data parallelism [10]. Unlike the "vanilla" version of Downpour SGD, our RL agents actively add experiences to the memory replay dataset (see Fig. 1).

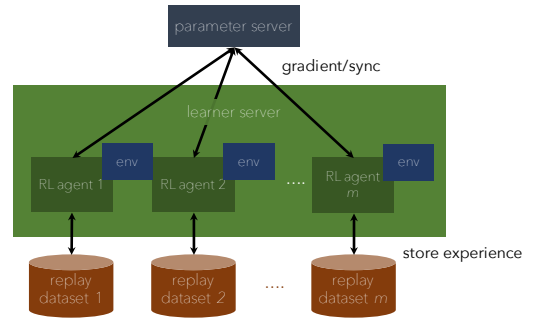

Fig. 1: A variant of Downpour SGD adapted to deep Q-learning.

A parameter server stores a global copy of the model. Each worker node is responsible for

1) fetching the latest model, $\theta$, from the server

**Algorithm 2** Distributed deep Q-learning

    **function** RMSPROPUPDATE($\Delta\theta$)
        $r_i \leftarrow 0.9 r_i + 0.1 (\Delta\theta)_i^2$ for all $i$
        Acquire write-lock
        $\theta_i \leftarrow \theta_i - \alpha (\Delta\theta)_i / \sqrt{r_i}$
        Release write-lock
    **Initialize server** $\theta_i \sim \mathcal{N}(0, \xi)$, $r \leftarrow 0$, $n \leftarrow 0$
    **for all** workers $k$ **do**
        **AsyncStart**
            Populate $\mathcal{D}_k$ by playing with random actions
            **repeat**
                COMPUTEGRADIENT
            **until** server timeout
        **AsyncEnd**
    **while** $n <$ MaxIters **do**
        **if** latest model requested by worker $k$ **then**
            Acquire write-lock
            **Send** $(\theta, n)$ to worker $k$
            Release write-lock
        **if** gradient $\Delta\theta$ received from worker $k$ **then**
            RMSPROPUPDATE($\Delta\theta$)
            $n \leftarrow n + 1$
    Shutdown server

2) generating new data for its local shard
3) computing a gradient using this model and mini-batch from its local replay memory dataset
4) sending the gradient $\Delta\theta$ back to the server.

These operations constitute a single worker iteration. All workers perform these iterations independently, asynchronously fetching from and pushing to the parameter server. Upon receiving a gradient, the parameter server *immediately* applies an update to the global model. The only synchronization is a write-lock on the model as it is being written to the network buffer.

For typical sizes of *Q*-networks, it takes much longer to compute a mini-batch gradient than it does to perform a parameter update. Therefore we can train on more data in the same amount of time by simply adding worker nodes (eventually this breaks, as we discuss later). Since each mini-batch is supposed to be drawn uniformly at random from some history of experiences, we can keep completely independent histories on each of the worker nodes and sample only from the local dataset. This allows our algorithm to scale extremely well with the size of the training data.

### B. Model parallelism

Google's implementation of Downpour SGD distributes each model replica across multiple machines. This allows them to scale to very large models. Our implementation, on the other hand, assumes that the model fits on a single machine. This places a strict upper bound on the size of the model. However, it allows us to easily take advantage of hardware acceleration for a single node's computation. We use the Caffe deep learning framework to perform the gradient computation on each of the worker machines. Caffe allows us to take advantage of fast BLAS implementations, using the worker's CPU or GPU resources.

In this sense, the work done by a single node is also parallelized—either across multiple CPU cores or many GPU cores. Pushing the computation down to the GPU yields a substantial speed up, but it further constrains the size of the model. The GPU memory must not only hold the model and batch of data, but also all the intermediate outputs of the feedforward computation. This memory limit is often approached in practice, especially for lower end GPUs. The CPU's memory is much more accommodating and should be able to hold any reasonably large model. In the case where the worker computation *must* be done on the CPU due to memory constraints, the advantages of distributed deep Q are even more drastic.

### C. Communication pattern

The server must communicate with all workers since each requires the latest model at every iteration. Each worker, in turn, communicates with the server to send a gradient, but does not need to communicate with any other worker node. This is similar to a one-to-all and all-to-one communication pattern, which could benefit from the `allreduce` primitive. However, all of the communication happens asynchronously, breaking the `allreduce` communication pattern. Further, to minimize the "staleness" of the model used by a worker for its gradient computation, it should fetch the latest version of the model *directly* from the server, not in bit-torrent fashion from its peers.

### D. Scalability issues

As we scale up the number of worker nodes, certain issues become increasingly important in thinking about the performance of the distributed deep Q-learning algorithm.

*1) Server bottleneck:* With few machines, the speed of training is bound by the gradient computation time. In this regime, the frequency of parameter updates grows linearly with the number of workers. However, the server takes some finite amount of time, $\tau$, to receive a gradient message and apply a parameter update. Thus, even with an infinite pool of workers, we cannot perform more than $1/\tau$ updates per second.

This latency, $\tau$, is generally small compared to the gradient computation time, but it becomes more significant as the number of workers increases. Suppose a mini-batch gradient can be computed in time $T$. A pool of $P$ workers will—on average—serve up a new gradient every $T/P$. Thus, once we have $P = T/\tau$ workers, we will no longer see any improvement by adding nodes.

This is potentially alarming, especially since both $T$ and $\tau$ grow linearly with the model size (i.e. the ratio $T/\tau$ is constant). However, one way to improve performance beyond this limit is to increase the batch size. Note that this increases the single worker computation time $T$, but does not affect the server latency $\tau$. Another option is to use a powerful machine for the server and further optimize our server-side code to reduce $\tau$.

*2) Gradient staleness:* As the frequency of parameter updates grows, it becomes increasingly likely that a gradient received by the server was computed using a significantly outdated model. This increases the noise in the parameter updates and could potentially slow down the optimization process and lead to problems with convergence. In practice, using adaptive learning rate methods like RMSprop or AdaGrad, we do not see such issues. However, as the number of workers increases, this could become a significant problem and should be examined carefully.

*E. Implementation details*

We rely on a slighted out-of-date version of Caffe (which is included as a submodule) as well as Spark for running the multi-worker version of distributed deep Q-learning. Our implementation does not make heavy use of Spark. However, Spark does facilitate scheduling of gradient updates, coordinating the addresses of all machines involved in the computation, and shipping the necessary files and serialized code to all of the worker nodes. We also made some progress towards a more generic interface between Caffe and Spark using `MemoryDataLayer`s and shared memory. For this, please see the `shmem` branch of the GitHub repository.

*F. Complexity analysis*

*1) Convolutional neural network:* To analyze our model complexity, we break our neural network into three components. The first part consists of the convolutional layers of the convolutional neural network (CNN). The complexity for this part is $O\left(d^2 F k^2 N^2 L_C\right)$, where we have frame width $d$, frame count $F$, filter width $k$, filter count $N$, convolution layer count $L_C$. The second part consists of the fully connected layers and its complexity is $O\left(H^2 L_B\right)$, where $H$ is the node count and $L_B$ is the hidden layer count. Finally, the "bridge" between the convolutional layers and fully connected layers contributes a complexity of $O\left(H d^2 N\right)$. The total number of parameters in the model $p$ is thus $O\left(d^2 F k^2 N^2 L_C + H^2 L_B + H d^2 N\right)$. We use the variable $p$ to simplify our notation.

*2) Runtime:* Consider a single worker and its interaction with the parameter server. The run-time for a single parameter update is the time to compute a gradient, $T$, plus the time to perform the update, $\tau$. Both $T$ and $\tau$ are $O(p)$, but the constant factor differs substantially. Further, the server takes at least time $\tau$ to perform an update, regardless of the number of workers.

Thus the run-time for $N$ parameter updates using $k$ worker machines is $O(Np/k)$ if $k < T/\tau$ or $O(N\tau)$ otherwise.

*3) Communication cost:* Each iteration requires both a model and a gradient to be sent over the network. This is $O(p)$ data. We do this for $N$ iterations. Thus the total communication cost is $O(Np)$.

## V. NUMERICAL EXPERIMENTS

To validate our approach, we apply it on the classic game of Snake and empirically demonstrate the performance of our algorithm.

*A. Snake*

We implemented the Snake game in Python. To generate experiences, our program preprocesses the frames of the game and feeds them into our neural network (see Fig. 2). In our implementation, the Spark driver sends a script to each worker that spawns a local game instance and a neural network. Using the neural network, the worker generates experience tuples by interacting with the game. The game is made up of an $n \times n$ array. The snake starts with body length of two and gains an additional body length when it eats an "apple." Ingesting an apple awards the agent one point. At game termination, conditioned on the snake hitting itself, the agent loses one point. The goal of the game is to maximize the score by having the snake eat more apples and not dying. Each worker sends their model to the server after every gradient computation, and receives the latest model from the server periodically as detailed in Section IV.

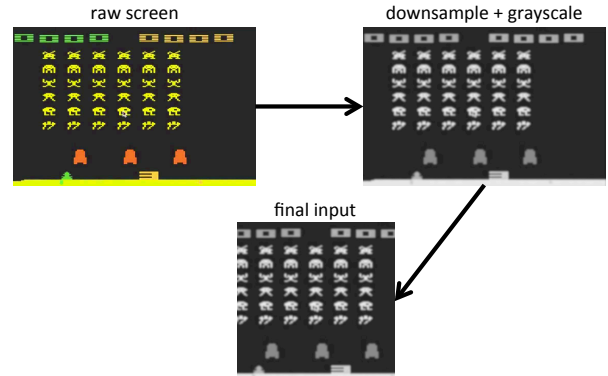

Fig. 2: Preprocessing consists of downsampling and grayscaling the image, and then cropping the image to the input size of our neural network.

*B. Computation and communication*

Figure 3 shows the experiment runtimes with different model sizes, which correspond to different game frame sizes. The legend is as follows. "comms" refers to the amount of time required for sending the model parameters between (either way) the parameter server and one worker. "gradient" refers to the compute time required to calculate a gradient update by a worker. "latency" refers to the amount of time required by the parameter server to update its weights with one set of gradients. The training rate was compute-bound by gradient calculations by each worker in our experiments.

Note that the gradient calculation line is two orders of magnitude larger than the other two lines in the figure. Note also that the upper bound on number of updates per second is inversely proportional to number of parameters in the model, since the single parameter server cannot update its weights faster than a linear rate with respect to the number of updates in a serial fashion. Thus we observe that as the number of workers and model size increase, the update latency could become the bottleneck of the learning process. To prevent such bottlenecks, we can increase the minibatch size for each gradient update. This modification would therefore increase the compute time required by each worker machine and therefore reduce the rate at which gradient updates are sent to the parameter server. Additionally, the gradient estimates would be better due to the larger minibatch size.

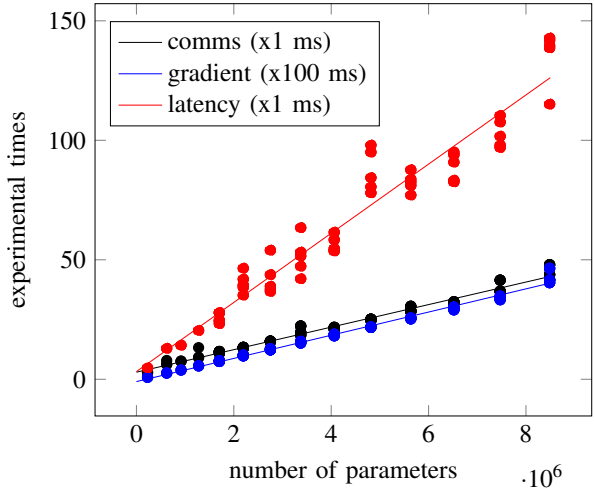

Fig. 3: Compute and communication times for various processes.

## C. Distributed Performance

To validate our work we collected results from two experiments, which were ran for a total of 120,000 gradient updates. The first experiment was performed using a serial implementation as documented in [4], [5]. The second experiment was ran with two workers for the same number of updates. As shown in Fig. 4, the two workers model experienced a much faster learning rate than the single worker model. In fact, the average reward over time scales linearly with number of workers: At every time stamp, we see that the average reward obtained by the two worker experiment is roughly twice of the single worker experiment. This trend suggests that the performance of our distributed algorithm scales linearly in the initial training phase.

## VI. CONCLUSION AND FUTURE WORK

We have developed a distributed deep Q-learning algorithm that can efficiently train an complicated RL agent on multiple machines in a cluster. The algorithm combines the sequential deep Q-learning algorithm developed in [4], [5] with DistBelief [10], accelerating the training process via asynchronous gradient updates from multiple machines at a linear rate of increase with the number of RL agents. Future work includes scaling up the experiments and studying how issues such as model staleness from having more worker machines impacts convergence rates. We will also compare our work with Gorila, a distributed deep learning architecture similar to ours that is pre-dated by the first version of our paper [17].

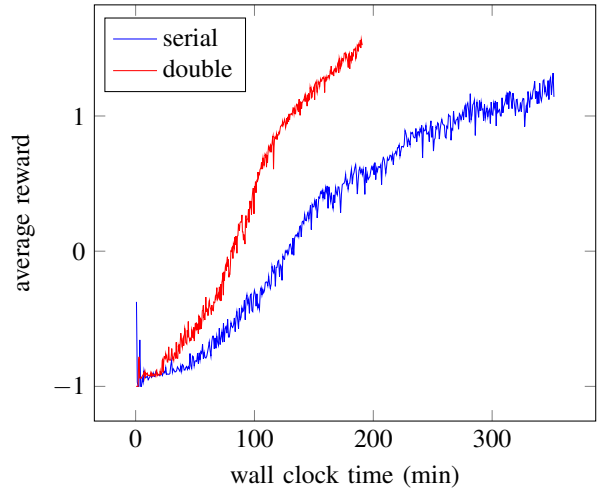

Fig. 4: Comparison of average reward increase between the serial and distributed implementations. The distributed implementation used two workers.

### SUPPLEMENTARY MATERIAL

All code for this project can be found, together with documentation, at

`https://github.com/kjchavez/distributed-deep-q`.

APPENDIX

*A. Network architecture*

Figure 5 visualizes the exact architecture of our deep neural network, as implemented in Caffe. Note that unlike in the "vanilla" DQN agent developed in [4], [5], our variant was designed as a single RL agent with both the target and actively training neural networks. This feature was enabled by special layers provided in the Caffe framework.

Fig. 5: Deep neural network architecture for numerical experiments.